\pdfoutput=1

\PassOptionsToPackage{table}{xcolor}

\documentclass[11pt]{article}

\usepackage[final]{acl2025}

\usepackage{latexsym}

\usepackage{multirow}
\usepackage{graphicx}
\usepackage{array}
\usepackage{tabularx}
\usepackage{lipsum}
\usepackage{arydshln}
\usepackage{paralist}
\usepackage{booktabs}  
\usepackage{placeins} 

\usepackage{xfp}
\definecolor{HeatAcc}{RGB}{33,102,172}   
\definecolor{HeatTime}{RGB}{150,111,71}  
\definecolor{HeatCall}{RGB}{136,86,167}  
\newcommand{\hc}[5]{\cellcolor{#1!\fpeval{round(15+57*max(0,min(1,(#4-#2)/(#3-#2))),0)}}#5}

\usepackage{tikz}
\usetikzlibrary{arrows.meta,positioning,fit,backgrounds,calc,shapes.geometric}
\usepackage{fontawesome5}  

\usepackage{times}
\usepackage{latexsym}

\usepackage{amsmath}
\usepackage{amssymb}
\usepackage{amsfonts}

\usepackage[T1]{fontenc}

\usepackage[utf8]{inputenc}

\usepackage{microtype}

\usepackage{inconsolata}

\title{\textsc{Ascent}: An Agentic System over the Model Context Protocol for Real-World Clinical Data Analysis}

\author{
  Angelo Ziletti\textsuperscript{*} \quad Leonardo D'Ambrosi \quad Melanie Tuchardt \quad Tim Kondziella \\
  Bayer AG, Germany \\
  \texttt{*angelo.ziletti@bayer.com}
}

\begin{document}
\maketitle

\begin{abstract}
Answering epidemiological questions from real-world clinical data requires
medical coding, schema-aware SQL, and validation of implicit choices about
populations, denominators, and time. We present \textsc{Ascent}, an agentic
system that exposes medical coding, question answering, and cohort analysis
through a shared Model Context Protocol tool surface for standardized and
native schemas. We introduce \textsc{EpiTrap}, a dataset testing whether systems
avoid recognized pharmacoepidemiological errors, and compare a fixed pipeline
with agents across models and orchestrators. With capable models, agents improve
accuracy over the fixed pipeline by an average of 27 and 20 percentage points on
native and standardized schemas, respectively. These gains require more tool calls and
longer runtimes. Experience from real projects highlights the system's value for
feasibility assessment, diagnostic iteration, and expert-guided analysis.
\end{abstract}

\section{Introduction} \label{sec:intro}

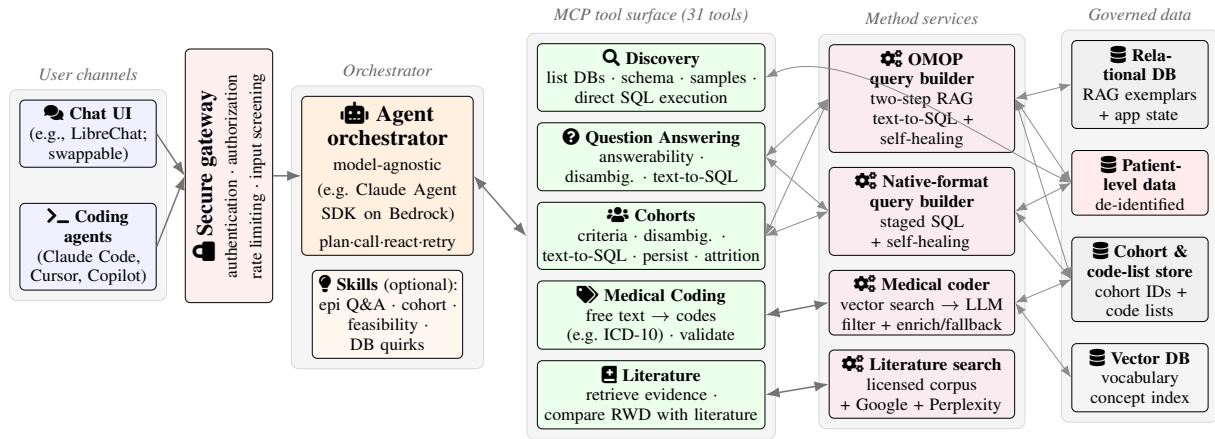
\begin{figure*}[t]
  \centering
  \resizebox{\textwidth}{!}{%
  \begin{tikzpicture}[
    font=\small,
    >=Latex,
    box/.style={draw, rounded corners=2pt, align=center, inner sep=2pt, minimum height=6mm},
    client/.style={box, fill=blue!6, text width=18mm, font=\scriptsize},
    gate/.style={box, fill=red!6, text width=6mm, font=\scriptsize},
    agent/.style={box, fill=orange!12, text width=23mm, minimum height=9mm},
    tool/.style={box, fill=green!8, text width=31mm, font=\scriptsize},
    engine/.style={box, fill=purple!8, text width=25mm, font=\scriptsize},
    store/.style={box, fill=gray!12, text width=18mm, font=\scriptsize},
    grouplbl/.style={font=\scriptsize\itshape, text=black!60},
    flow/.style={->, semithick, black!55},
    mesh/.style={->, thin, black!45},
  ]

  \node[client] (webui) at (0,1.5) {\faComments~\textbf{Chat UI}\\{\scriptsize (e.g., LibreChat;}\\{\scriptsize swappable)}};
  \node[client, below=5mm of webui] (agents) {\faTerminal~\textbf{\mbox{Coding} agents}\\{\scriptsize (Claude Code,}\\{\scriptsize Cursor, Copilot)}};

  \node[box, fill=red!6, right=4mm of webui.east, yshift=-6mm, minimum height=36mm,
        inner sep=3pt] (gw)
    {\rotatebox{90}{\parbox{34mm}{\centering
       \faLock~\textbf{Secure gateway}\\[2pt]
       {\scriptsize authentication $\cdot$ authorization\\
        rate limiting $\cdot$ input screening}}}};

  \node[agent, right=4mm of gw.east] (agent)
    {\faRobot~\textbf{Agent orchestrator}\\{\scriptsize model-agnostic}\\{\scriptsize (e.g.\ Claude Agent}\\{\scriptsize SDK on Bedrock)}\\[1pt]{\scriptsize plan·call·react·retry}};
  \node[box, fill=orange!6, below=2mm of agent, text width=20mm, font=\scriptsize] (skills)
    {\faLightbulb~\textbf{Skills} (optional):\\ epi Q\&A · cohort ·\\ feasibility · DB quirks};

  \node[tool, right=9mm of agent.east, yshift=14mm] (discovery)
    {\faSearch~\textbf{Discovery}\\ list DBs · schema · samples ·\\ direct SQL execution};
  \node[tool, below=1.5mm of discovery] (qa)
    {\faQuestionCircle~\textbf{Question Answering}\\ answerability ·\\ disambig. · text-to-SQL};
  \node[tool, below=1.5mm of qa] (cohort)
    {\faUsers~\textbf{Cohorts}\\ criteria · disambig. ·\\ text-to-SQL · persist · attrition};
  \node[tool, below=1.5mm of cohort] (coding)
    {\faTags~\textbf{Medical Coding}\\ free text $\to$ codes\\ (e.g.\ ICD-10) · validate};
  \node[tool, below=1.5mm of coding] (lit)
    {\faBookMedical~\textbf{Literature}\\ retrieve evidence ·\\ compare RWD with literature};

  \node[engine, right=9mm of discovery.east, yshift=-3mm] (omop)
    {\faCogs~\textbf{OMOP query builder}\\ two-step RAG\\ text-to-SQL + self-healing};
  \node[engine, below=2mm of omop] (native)
    {\faCogs~\textbf{Native-format query builder}\\ staged SQL\\ + self-healing};
  \node[engine, below=2mm of native] (coder)
    {\faCogs~\textbf{Medical coder}\\ vector search $\to$ LLM\\ filter + enrich/fallback};
  \node[engine, below=2mm of coder] (litsvc)
    {\faCogs~\textbf{Literature search}\\ licensed corpus\\ + Google + Perplexity};

  \node[store, right=8mm of omop.east, yshift=2mm] (pg)
    {\faDatabase~\textbf{Relational DB}\\ RAG exemplars\\ + app state};
  \node[store, below=3mm of pg, fill=red!8] (snow)
    {\faDatabase~\textbf{Patient-level data}\\ de-identified};
  \node[store, below=3mm of snow] (cohortstore)
    {\faDatabase~\textbf{Cohort \& code-list store}\\ cohort IDs +\\ code lists};
  \node[store, below=3mm of cohortstore, fill=gray!8] (qdrant)
    {\faDatabase~\textbf{Vector DB}\\ vocabulary\\ concept index};

  \begin{scope}[on background layer]
    \tikzset{band/.style={draw=black!15, fill=black!4, rounded corners=3pt, inner sep=4pt}}
    \node[band, fit=(webui)(agents), label={[grouplbl]above:User channels}] {};
    \node[band, fit=(agent)(skills), label={[grouplbl]above:Orchestrator}] {};
    \node[band, fit=(discovery)(cohort)(qa)(coding)(lit),
          label={[grouplbl]above:MCP tool surface (31 tools)}] (mcp) {};
    \node[band, fit=(omop)(native)(coder)(litsvc), label={[grouplbl]above:Method services}] {};
    \node[band, fit=(pg)(snow)(cohortstore)(qdrant), label={[grouplbl]above:Governed data}] {};
  \end{scope}

  \draw[flow] (webui.east)  -- (gw.west);
  \draw[flow] (agents.east) -- (gw.west);
  \draw[flow] (gw.east) -- (agent.west);
  \draw[flow, <->] (agent.east) -- (mcp.west);

  \draw[mesh, <->] (cohort.east)    -- (omop.west);
  \draw[mesh, <->] (cohort.east)    -- (native.west);
  \draw[mesh, <->] (qa.east)        -- (omop.west);
  \draw[mesh, <->] (qa.east)        -- (native.west);
  \draw[flow, <->] (coding.east) -- (coder.west);
  \draw[flow, <->] (lit.east)   -- (litsvc.west);
  \draw[mesh, <->] (discovery.east) to[out=25,in=155,looseness=0.7] (snow.west);

  \draw[mesh, <->] (omop.east)   -- (pg.west);
  \draw[mesh, <->] (omop.east)   -- (snow.west);
  \draw[mesh, <->] (omop.east)   -- (cohortstore.west);
  \draw[mesh, <->] (native.east) -- (snow.west);
  \draw[mesh, <->] (native.east) -- (cohortstore.west);
  \draw[mesh, <->] (coder.east)  -- (qdrant.west);
  \draw[mesh, <->] (coder.east)  -- (cohortstore.west);

  \end{tikzpicture}%
  }
  \caption{\textbf{\textsc{Ascent} architecture.} MCP clients and model-agnostic
  orchestrators access shared analytical services and governed data stores through
  a common tool surface.}
  \label{fig:architecture}
\end{figure*}

Answering epidemiological questions from real-world clinical data requires more
than translating language into SQL. Analysts must define populations and time
windows, map clinical concepts to codes, account for heterogeneous schemas, and
verify that an executable query answers the intended question.

Existing conversational and text-to-SQL systems demonstrate the potential to
reduce the implementation burden of clinical data analysis~\citep{boyle-etal-2026-clinqueryagent,ziletti-dambrosi-2024-retrieval,marshan-etal-2024-medt5sql}.
However, epidemiological questions often leave consequential choices about
population, denominator, and time window implicit, allowing an executable query
to return a plausible answer to the wrong question. An agentic system can surface
these choices, inspect intermediate results, and revise the analysis before
reporting.

We present \textsc{Ascent}, an agentic system deployed at Bayer
for expert-supervised analysis of real-world clinical data. Rather than embedding
analytical capabilities in one application or agent loop, \textsc{Ascent} exposes
them through a shared Model Context Protocol (MCP)~\citep{anthropic-mcp-2024} tool
surface. This separates the governed data-access layer from the client, model,
and orchestrator while supporting both the standardized Observational Medical
Outcomes Partnership Common Data Model (OMOP-CDM)~\citep{omop-cdm-2023} and
native vendor schemas.

\paragraph{Contributions.}
\begin{compactitem}
  \item \textsc{Ascent}, a deployed, governed MCP tool surface that decouples
        medical coding, epidemiological analysis, and cohort building from the
        client, model, and orchestrator across OMOP and native data.
  \item \textsc{EpiTrap}, a parallel native-schema and OMOP evaluation set
        designed around recognized epidemiological errors.
  \item An evaluation across models and orchestrators, together with lessons from
        real-world use.
\end{compactitem}
\noindent We publicly release both the \textsc{Ascent} code and the
\textsc{EpiTrap} dataset.\footnote{\url{https://github.com/bayer-group/ascent-epi-demo}}

\section{Related Work} \label{sec:related}

\paragraph{Clinical conversational agents and MCP-native systems.}
Our closest neighbor is \textsc{ClinQueryAgent}~\citep{boyle-etal-2026-clinqueryagent},
a task-oriented dialogue agent, trialled with NHS staff, that turns
population-health questions into a logical query language over a curated concept
library, delegating concept retrieval to a sub-agent to control context growth.
\textsc{Ascent} differs in two key respects: it exposes its capabilities as a
portable, model-agnostic MCP tool surface that any client can orchestrate freely,
rather than a single hand-crafted agent loop, and it generates open SQL directly
over \emph{both} OMOP and native schemas. Broader clinical text-to-SQL work spans agentic
querying, cohort reasoning, and multi-turn evaluation, while continuing to expose
challenges in interaction robustness and clinical-code
synthesis~\citep{lee2022ehrsql,shi-etal-2024-ehragent,ec2seq2sql,shen-etal-2026-patient,ehrchatqa2025,codeclinic2026}.
In parallel, clinical systems increasingly use MCP to connect agents with data
and tools~\citep{ehrmcp2025,frei-etal-2026-infherno,stmcp2026}.

\paragraph{RWD platforms and general-purpose text-to-SQL.}
The open-source OHDSI stack (\textsc{Atlas}, \textsc{Hades})~\citep{ohdsi-atlas,ohdsi-hades}
provides rule-based OMOP cohort definition, with natural-language support
explored separately by systems such as
\textsc{Criteria2Query}~\citep{criteria2query,park-etal-2024-criteria2query3}.
Commercial RWD platforms such as TriNetX~\citep{trinetx},
Aetion~\citep{aetion-ai}, and Truveta~\citep{truveta-intelligence} provide
integrated analytics but are closed, platform-specific environments. Conversely,
warehouse-native systems such as \textsc{Snowflake Cortex
Analyst}~\citep{cortex-analyst} and \textsc{Databricks Genie}~\citep{databricks-genie}
generate SQL but lack the medical coding and epidemiological semantics required
for RWD analysis.

\paragraph{Ambiguity and reflection.}
Epidemiological questions are often underspecified, motivating clarification
before analysis~\citep{ziletti-etal-2026-disentangling,
liu-etal-2026-mind-ambiguity}. Whether agents benefit from reflection remains
unsettled: it can improve agentic performance~\citep{shinn2023reflexion}, but
models may overlook their own errors or degrade correct
answers~\citep{huang2023selfcorrect,kamoi-etal-2024-llms}. Clinical agents have
therefore explored structured reflection and critique--revision
workflows~\citep{liao-etal-2025-reflectool,ghafoor-etal-2025-medical-safety}.

\section{System Overview} \label{sec:system}

\textsc{Ascent} comprises a client-facing agent, an MCP tool surface, and shared
method services backed by dedicated data stores
(Fig.~\ref{fig:architecture}). The model selects and chains tools into an
analysis. We next describe the tool surface, its supporting services, and
deployment. Fig.~\ref{fig:trace} shows an interaction through a chat interface.

\subsection{Agent and MCP tool surface}
The orchestrator is a large language model (LLM) agent that plans, issues tool calls, inspects
intermediate results, and retries or re-plans. Its only interface to data is an
MCP server exposing 31 tools.\footnote{The full tool list is available
in the code release: \url{https://github.com/bayer-group/ascent-epi-demo}.}
\textsc{Ascent} exposes its capabilities through this shared MCP tool surface
rather than a bespoke agent loop. This decouples the client, orchestration
runtime, model, and governed data-access layer. The design provides three
properties. (i)~\emph{Portability}: the same tool surface supports a self-hosted
chat front-end (e.g., LibreChat~\citep{librechat}), coding agents such as Claude Code
and Cursor, and other MCP-compatible clients. (ii)~\emph{Model-agnosticism}: the
orchestrating model or agent can be replaced without modifying the tools.
(iii)~\emph{Isolation}: the model accesses data only through governed tools.

The tools are organized into five capability groups that reuse four shared
back-end services (\S\ref{sec:backend}).

\noindent\textbf{Discovery.} Enumerate the databases a user may access, return
machine-readable schemas, list supported ontologies, and surface datasource
metadata, row counts, and sample rows.

\noindent\textbf{Medical coding.} Translate free-text concepts into standardized
codes (e.g.\ resolving ``ischemic stroke'' to its ICD-10 and other vocabulary
codes), validate them against a vocabulary, and report per-code patient counts to
show which codes are actually used in the data.

\noindent\textbf{Question answering.} Take an epidemiological question and produce
the SQL that answers it, together with its results. A lightweight
\emph{answerability} check first estimates, against the \textsc{EpiAskKB}
retrieval corpus~\citep{ziletti-etal-2025-cohort}, whether the question is
plausibly supported. Underspecified questions are disambiguated with the
\textsc{Clues} method~\citep{ziletti-etal-2026-disentangling}, which surfaces the
consequential interpretations of a question for the user to confirm or edit. The
disambiguated question is then translated to SQL using retrieval-augmented
text-to-SQL for OMOP~\citep{ziletti-dambrosi-2024-retrieval} or schema-grounded
generation for native schemas (\S\ref{sec:nonomop}).

\noindent\textbf{Cohorts.} Build patient cohorts from natural-language
eligibility criteria by structuring the criteria and generating SQL using
retrieval-augmented cohort generation for OMOP~\citep{ziletti-etal-2025-cohort}
or schema-grounded generation for native schemas (\S\ref{sec:nonomop}). Cohorts
can be persisted and shared, with attrition funnels and baseline summaries of
demographics, comorbidities, medications, and labs.

\noindent\textbf{Literature.} Retrieve scientific evidence for a question and
compare a real-world-data answer against published literature.

Direct SQL execution complements these structured workflows, letting the agent
inspect unfamiliar tables, verify denominators, and repair mis-scoped filters,
then re-query. Combining the curated tools with this open-ended access lets
analytical operations be selected, inspected, and revised dynamically.

\begin{figure}[h!tb]
  \centering
  \includegraphics[width=\columnwidth]{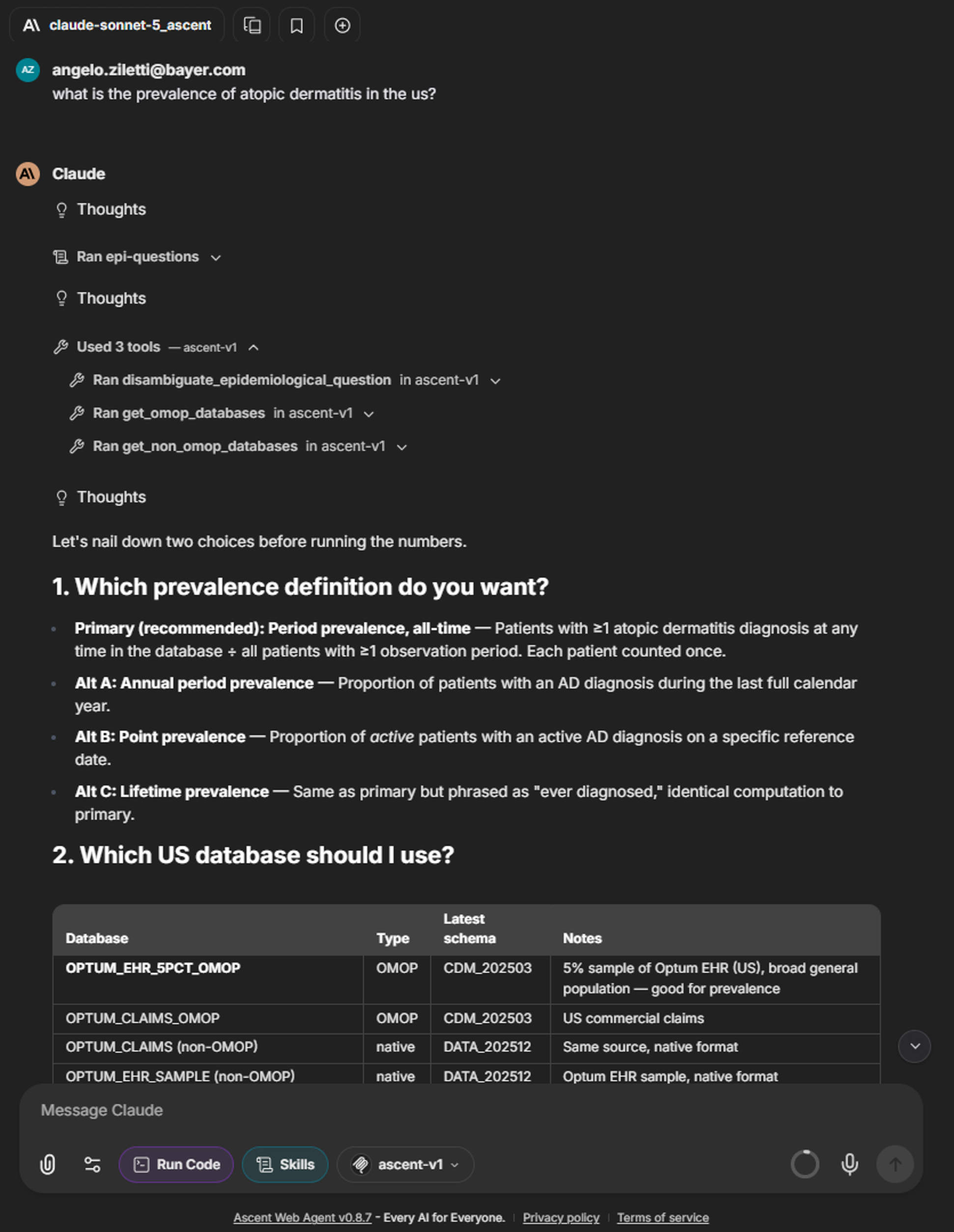}
  \caption{\textbf{Example \textsc{Ascent} session through a chat interface.}
  After loading an appropriate skill, the system surfaces consequential choices
  about the prevalence definition and target database.}
  \label{fig:trace}
\end{figure}

\paragraph{Skills.}
Optional, on-demand \emph{skills} encode vetted analytical workflows, feasibility
checks, and database-specific conventions, such as coding formats, identifier
capitalization, and date semantics. They guide, rather than replace, the
orchestrator's reasoning over the tools (Fig.~\ref{fig:trace}).

\begin{table*}[t]
  \centering
  \small
  \renewcommand{\arraystretch}{1.15}
  \begin{tabularx}{\textwidth}{@{}X X@{}}
    \toprule
    Question (system sees only this) & The recognized error the reasoning must avoid \\
    \midrule
    \emph{Count patients with severe hyperglycemia (blood glucose above 250~mg/dL) at any measurement.}
       & \textbf{Unit harmonization.} Glucose is recorded in mixed units, so applying one threshold to raw values misclassifies measurements. Correct: harmonize units, remove implausible values, then apply the threshold. \\
    \addlinespace[2pt]
    \emph{Estimate the one-year incidence of gout among adults with at least one year of continuous enrollment.}
       & \textbf{Prevalence--incidence conflation.} A pre-existing gout diagnosis is not an incident case. Correct: exclude prevalent cases via a baseline washout. \\
    \addlinespace[2pt]
    \emph{Measure the cervical-cancer screening rate among adult women.}
       & \textbf{Denominator ineligibility.} Women with documented absence of the cervix (e.g.\ following total hysterectomy) are ineligible for routine screening. Correct: exclude them from the denominator using prior diagnoses and procedures. \\
    \bottomrule
  \end{tabularx}
  \caption{\textbf{Representative \textsc{EpiTrap} questions}, spanning
  data-quality, estimand-definition, and denominator-eligibility errors. The
  system sees only the left column; it passes if its reasoning avoids the error
  in the right, whatever path it takes.}
  \label{tab:benchmark-examples}
\end{table*}

\subsection{Method services}
\label{sec:backend}
Four shared services implement the tools: OMOP and native-schema query builders,
a medical coder, and literature retrieval. In the evaluated deployment, OMOP and
native-schema SQL generation use Opus~4.8~\citep{anthropic_opus48_2026} and
Gemini~3.6 Flash~\citep{googledeepmind2026gemini36flash}, respectively.
Both query builders serve question answering and cohort generation, with
\textsc{Clues} disambiguation applied upstream and the medical coder used to
resolve clinical concepts. The OMOP builder
follows prior retrieval-augmented
methods~\citep{ziletti-dambrosi-2024-retrieval,ziletti-etal-2025-cohort}; we
summarize below the native builder, new to this paper, and the medical coder.

\subsubsection{Native-format query builder}
\label{sec:nonomop}
Native claims and EHR schemas differ in tables, columns, and coding conventions.
The builder first retrieves an \emph{M-Schema} representation containing tables,
columns, types, and sampled values~\citep{xiyansql,xiyansql_pre}, together with
column value distributions and a map of which columns carry ontology-coded values
for the medical coder. A model then
extracts the population, temporal frame, and analytical steps and generates SQL
with placeholders for coded concepts. The medical coder resolves these
placeholders using source-appropriate normalization, such as dot-stripped ICD
codes or padded NDCs. The query is executed on Snowflake and repaired
automatically following execution errors.

\subsubsection{Medical coder}
\label{sec:coder}
The coder retrieves candidate vocabulary concepts by dense search using a
clinical encoder~\citep{liu-etal-2021-self,bge_embedding,biolord2024}, after which
an LLM filters candidates for clinical consistency. Domain-specific expansion adds
relevant descendants, drug-class members, or NDC variants. Patient counts indicate
which codes occur in the target data, and cached, provenance-tracked code lists
can be reused across analyses.

\subsection{Deployment and persistence}
\textsc{Ascent} runs as a FastAPI backend with persistence separated by function
(Fig.~\ref{fig:architecture}): patient-level data are stored in Snowflake, while
RAG exemplars, vocabulary indexes, and code lists are stored separately. MCP
propagates the caller's identity so queries execute under the user's Snowflake
grants and are logged; the underlying patient data are de-identified and
accessed within the governed enterprise boundary.

\section{\textsc{EpiTrap}: An Epidemiological Dataset of Designed-In Reasoning Errors}
\label{sec:benchmark}

\noindent\textbf{Why a new dataset.} Existing clinical text-to-SQL datasets
primarily assess single-shot SQL against reference queries or execution
results~\citep{ziletti-dambrosi-2024-retrieval,legrand-etal-2026-omop}. This
approach does not fully capture agents that inspect schemas, run diagnostics,
and revise definitions before answering. Reference SQL privileges one solution
path, while enumerating all valid trajectories is impractical and does not
directly test epidemiological reasoning.

\noindent\textbf{Error-first evaluation.}
Each question targets one designed-in error and has a trap-specific rubric
describing the naive approach that produces the error and the requirement a
correct query must satisfy to avoid it. A response passes if its logic avoids
the designated error, regardless of its tool calls or SQL revisions; a run that
produces no executed query counts as a failure. This path-independent criterion
applies to both fixed pipelines and multi-step agents and requires no numerical
answer key. Because each rubric tests only its designated error, accuracy is an
upper bound on end-to-end correctness. Table~\ref{tab:benchmark-examples} gives
representative examples.

\noindent\textbf{Error selection and grounding.}
We identified candidate errors through LLM-assisted literature search; questions
and rubrics were LLM-drafted and author-edited. Errors cover outcome
ascertainment, eligibility and denominators, temporal definitions, occurrence
measures, missingness, unit harmonization, and protopathic
bias~\citep{ray2003newuser,rothman2008modern,horwitz1980protopathic,kahn2016dataquality}.

\noindent\textbf{Multi-judge grading.}
Following the LLM-as-judge paradigm~\citep{zheng2023judging}, Claude
Opus~5~\citep{anthropic_opus5_2026} and GPT-5.6
Luna~\citep{openai_gpt56_2026} independently grade each response from its
question, rubric, SQL, and reasoning to limit single-model bias. Agreement is
82\% ($\kappa=0.62$) on native and 87\% ($\kappa=0.75$) on OMOP. Claude
Opus~4.8 adjudicates disagreements using both
judges' reasoning and the executed SQL, citing the deciding rubric clause.

\noindent\textbf{Two data models.} We instantiate parallel versions for the
native Optum\textsuperscript{\textregistered} de-identified EHR (US hospitals and
clinics; 156 questions) and OMOP-CDM (153 questions; three native items are
OMOP-infeasible), both evaluated here.
Their error-specific rubrics make evaluation independent of a particular query
or execution path and applicable to other RWD text-to-SQL and agentic systems.
We release the questions and trap-specific rubrics for both versions.

\section{Evaluation} \label{sec:eval}
We compare a fixed pipeline with agentic configurations across models and
orchestration frameworks, examining accuracy and resource use.

\subsection{Experimental setup}

The \textbf{mechanical pipeline} generates SQL, resolves medical codes, executes
the query, and verbalizes the result in a fixed sequence without revision.
\textbf{Agentic} systems may inspect schemas, select tools,
execute diagnostics, and revise their SQL. We test Claude Agent
SDK~\citep{anthropic2026claudeagentsdk}, LangGraph~\citep{langchain2026langgraph},
and Strands~\citep{strandsagents2026sdk} with Sonnet~5~\citep{anthropic2026sonnet5},
DeepSeek-V4-Pro~\citep{deepseekai2026deepseekv4}, and
gpt-oss-120b~\citep{openai2025gptoss}. Claude Agent SDK, restricted to Anthropic
models, is tested only with Sonnet~5. Each configuration uses either a minimal
\textbf{plain} prompt or a \textbf{reflect} prompt directing it to critique the
interpretation, study design, and result.

All configurations were evaluated on both \textsc{EpiTrap} versions through the
same production MCP server using Promptfoo~\citep{promptfoo}. Responses were
graded using the multi-judge protocol described in \S\ref{sec:benchmark}. For
average accuracy changes over the mechanical pipeline, we obtain 95\% confidence
intervals (CIs) using a paired question-level bootstrap.

\subsection{Results}

\begin{table}[htbp]
  \centering
  \footnotesize
  \setlength{\tabcolsep}{3pt}
  \begin{tabular}{l ccc ccc}
    \toprule
    & \multicolumn{3}{c}{Native, $n{=}156$}
    & \multicolumn{3}{c}{OMOP, $n{=}153$} \\
    \cmidrule(lr){2-4}\cmidrule(lr){5-7}
    Configuration & Acc. & Time & Calls & Acc. & Time & Calls \\
    \midrule
    \multicolumn{7}{l}{\emph{Baseline: no agent loop, no framework}}\\
    mech.\ pipeline        & \hc{HeatAcc}{38}{83}{50}{50\%} & \hc{HeatTime}{197}{631}{231}{231\,s} & \hc{HeatCall}{3}{22}{3.0}{3.0}  & \hc{HeatAcc}{47}{84}{49}{49\%} & \hc{HeatTime}{177}{677}{180}{180\,s} & \hc{HeatCall}{3}{22.8}{3.0}{3.0} \\
    \midrule
    \multicolumn{7}{l}{\emph{Claude Agent SDK}}\\
    \quad Sonnet 5, plain   & \hc{HeatAcc}{38}{83}{75}{75\%} & \hc{HeatTime}{197}{631}{565}{565\,s} & \hc{HeatCall}{3}{22}{9.4}{9.4}  & \hc{HeatAcc}{47}{84}{75}{75\%} & \hc{HeatTime}{177}{677}{275}{275\,s} & \hc{HeatCall}{3}{22.8}{8.8}{8.8} \\
    \quad \enspace$\hookrightarrow$ no scaffold   & \hc{HeatAcc}{38}{83}{76}{76\%} & \hc{HeatTime}{197}{631}{197}{197\,s} & \hc{HeatCall}{3}{22}{16.8}{16.8} & \hc{HeatAcc}{47}{84}{84}{84\%} & \hc{HeatTime}{177}{677}{177}{177\,s} & \hc{HeatCall}{3}{22.8}{18.6}{18.6} \\
    \quad Sonnet 5, reflect & \hc{HeatAcc}{38}{83}{83}{83\%} & \hc{HeatTime}{197}{631}{631}{631\,s} & \hc{HeatCall}{3}{22}{14.6}{14.6} & \hc{HeatAcc}{47}{84}{82}{82\%} & \hc{HeatTime}{177}{677}{355}{355\,s} & \hc{HeatCall}{3}{22.8}{12.0}{12.0} \\
    \quad \enspace$\hookrightarrow$ no scaffold & \hc{HeatAcc}{38}{83}{81}{81\%} & \hc{HeatTime}{197}{631}{329}{329\,s} & \hc{HeatCall}{3}{22}{22.0}{22.0} & \hc{HeatAcc}{47}{84}{84}{84\%} & \hc{HeatTime}{177}{677}{290}{290\,s} & \hc{HeatCall}{3}{22.8}{22.8}{22.8} \\
    \midrule
    \multicolumn{7}{l}{\emph{LangGraph}}\\
    \quad Sonnet 5, plain   & \hc{HeatAcc}{38}{83}{79}{79\%} & \hc{HeatTime}{197}{631}{467}{467\,s} & \hc{HeatCall}{3}{22}{7.2}{7.2}  & \hc{HeatAcc}{47}{84}{71}{71\%} & \hc{HeatTime}{177}{677}{268}{268\,s} & \hc{HeatCall}{3}{22.8}{6.6}{6.6} \\
    \quad Sonnet 5, reflect & \hc{HeatAcc}{38}{83}{79}{79\%} & \hc{HeatTime}{197}{631}{491}{491\,s} & \hc{HeatCall}{3}{22}{7.7}{7.7}  & \hc{HeatAcc}{47}{84}{71}{71\%} & \hc{HeatTime}{177}{677}{461}{461\,s} & \hc{HeatCall}{3}{22.8}{8.4}{8.4} \\
    \quad DeepSeek, plain   & \hc{HeatAcc}{38}{83}{75}{75\%} & \hc{HeatTime}{197}{631}{536}{536\,s} & \hc{HeatCall}{3}{22}{6.4}{6.4}  & \hc{HeatAcc}{47}{84}{75}{75\%} & \hc{HeatTime}{177}{677}{496}{496\,s} & \hc{HeatCall}{3}{22.8}{6.3}{6.3} \\
    \quad DeepSeek, reflect & \hc{HeatAcc}{38}{83}{75}{75\%} & \hc{HeatTime}{197}{631}{556}{556\,s} & \hc{HeatCall}{3}{22}{8.4}{8.4}  & \hc{HeatAcc}{47}{84}{71}{71\%} & \hc{HeatTime}{177}{677}{677}{677\,s} & \hc{HeatCall}{3}{22.8}{9.1}{9.1} \\
    \quad gpt-oss, plain    & \hc{HeatAcc}{38}{83}{53}{53\%} & \hc{HeatTime}{197}{631}{465}{465\,s} & \hc{HeatCall}{3}{22}{3.8}{3.8}  & \hc{HeatAcc}{47}{84}{59}{59\%} & \hc{HeatTime}{177}{677}{225}{225\,s} & \hc{HeatCall}{3}{22.8}{4.3}{4.3} \\
    \quad gpt-oss, reflect  & \hc{HeatAcc}{38}{83}{54}{54\%} & \hc{HeatTime}{197}{631}{486}{486\,s} & \hc{HeatCall}{3}{22}{4.4}{4.4}  & \hc{HeatAcc}{47}{84}{61}{61\%} & \hc{HeatTime}{177}{677}{191}{191\,s} & \hc{HeatCall}{3}{22.8}{4.8}{4.8} \\
    \midrule
    \multicolumn{7}{l}{\emph{Strands}}\\
    \quad Sonnet 5, plain   & \hc{HeatAcc}{38}{83}{75}{75\%} & \hc{HeatTime}{197}{631}{434}{434\,s} & \hc{HeatCall}{3}{22}{5.9}{5.9}  & \hc{HeatAcc}{47}{84}{65}{65\%} & \hc{HeatTime}{177}{677}{350}{350\,s} & \hc{HeatCall}{3}{22.8}{6.1}{6.1} \\
    \quad Sonnet 5, reflect & \hc{HeatAcc}{38}{83}{79}{79\%} & \hc{HeatTime}{197}{631}{439}{439\,s} & \hc{HeatCall}{3}{22}{7.3}{7.3}  & \hc{HeatAcc}{47}{84}{62}{62\%} & \hc{HeatTime}{177}{677}{411}{411\,s} & \hc{HeatCall}{3}{22.8}{8.0}{8.0} \\
    \quad DeepSeek, plain   & \hc{HeatAcc}{38}{83}{75}{75\%} & \hc{HeatTime}{197}{631}{540}{540\,s} & \hc{HeatCall}{3}{22}{6.1}{6.1}  & \hc{HeatAcc}{47}{84}{57}{57\%} & \hc{HeatTime}{177}{677}{633}{633\,s} & \hc{HeatCall}{3}{22.8}{4.7}{4.7} \\
    \quad DeepSeek, reflect & \hc{HeatAcc}{38}{83}{72}{72\%} & \hc{HeatTime}{197}{631}{595}{595\,s} & \hc{HeatCall}{3}{22}{7.8}{7.8}  & \hc{HeatAcc}{47}{84}{66}{66\%} & \hc{HeatTime}{177}{677}{648}{648\,s} & \hc{HeatCall}{3}{22.8}{7.5}{7.5} \\
    \quad gpt-oss, plain    & \hc{HeatAcc}{38}{83}{38}{38\%} & \hc{HeatTime}{197}{631}{391}{391\,s} & \hc{HeatCall}{3}{22}{3.5}{3.5}  & \hc{HeatAcc}{47}{84}{47}{47\%} & \hc{HeatTime}{177}{677}{307}{307\,s} & \hc{HeatCall}{3}{22.8}{3.8}{3.8} \\
    \quad gpt-oss, reflect  & \hc{HeatAcc}{38}{83}{43}{43\%} & \hc{HeatTime}{197}{631}{411}{411\,s} & \hc{HeatCall}{3}{22}{3.9}{3.9}  & \hc{HeatAcc}{47}{84}{48}{48\%} & \hc{HeatTime}{177}{677}{285}{285\,s} & \hc{HeatCall}{3}{22.8}{4.1}{4.1} \\
    \bottomrule
  \end{tabular}
  \caption{\textbf{\textsc{EpiTrap} results across two data models,} native Optum
  EHR and OMOP-CDM. \emph{Acc.}\ is rubric-based accuracy; \emph{Time} and
  \emph{Calls} are per-question means. ``No scaffold'' rows withhold the
  specialized text-to-SQL tools, leaving the agent to write SQL itself.}
  \label{tab:eval}
\end{table}

\paragraph{Agentic tool use improves accuracy with capable models.}
Averaged across the ten Sonnet~5 and DeepSeek configurations
(Table~\ref{tab:eval}), agentic tool use improves accuracy over the mechanical
pipeline by 26.7 points on the native schema (95\% CI [18.7, 35.0]) and 20.4
points on OMOP (95\% CI [13.3, 27.5]). By contrast, gpt-oss stays at baseline
($-3.0$ [$-11.2$, $5.3$] native; $+4.6$ [$-2.0$, $11.3$] OMOP).

\paragraph{Reflection adds cost without consistent gains.}
It improves accuracy with the Claude Agent SDK ($75\!\to\!83$ native,
$75\!\to\!82$ OMOP) but not consistently elsewhere. On average, it increases
tool calls by 27\% and runtime by 13\%.

\paragraph{Specialized SQL tooling reduces tool calls without improving accuracy
for a frontier model.}
With the Claude Agent SDK and Sonnet~5, we withhold the specialized
text-to-SQL tools. The agent still uses the remaining \textsc{Ascent} tools to
inspect schemas, resolve codes, and write and execute SQL. Accuracy remains
broadly unchanged, tool calls increase by $1.5$--$2.1\times$, and runtime
decreases (Table~\ref{tab:eval}, ``no scaffold'').
Consistent with prior findings for stronger models~\citep{Schmidgall2026-hn},
the specialized SQL tooling provides fewer interaction steps and reusable,
governed artifacts, but no gains in accuracy or speed, a relevant trade-off in
clinical-agent evaluation~\citep{ruhrberg2026benchmark}.
\FloatBarrier 

\section{Real-World Usage and Lessons Learned} \label{sec:usage}
From 12 July to 21 September 2026, 36 users made 13{,}002 calls through LibreChat,
Cursor, Claude Code, and other clients. We summarize logged sessions and
researcher-authored use-case reports.

\paragraph{Observed workflows.}
Practitioners moved beyond query generation, using \textsc{Ascent} to assess
feasibility, challenge assumptions, and refine populations, codes, index dates,
time windows, and outcomes. The agent supported schema inspection, coding, SQL
generation, and diagnostics, while experts evaluated clinical and methodological
defensibility.
Staged feasibility checks of concept occurrence, covariate coverage, endpoint
observability, and follow-up informed whether to proceed, revise the design, use
a proxy, or abandon an unsupported analysis; a defensible negative finding was
itself valuable, preventing investment in an infeasible study.

Cross-database analyses often required different claims and EHR
implementations because of differences in coding, observation, and available
measurements. Divergent estimates prompted examination of eligibility,
measurement, and healthcare-contact patterns, revealing how each source
represents clinical practice.

\paragraph{Observed analytical failure modes.}
Real usage exposed failures that successful execution alone could not detect:
\begin{compactenum}
  \item \textbf{Plausible answers to the wrong question.} A query could use the
        wrong population, vocabulary, index date, unit, or counting grain
        (e.g.\ counting individual billing records rather than distinct
        hospitalizations) while still returning a plausible result.
  \item \textbf{Silent scope drift in clinical definitions.} Automated code
        expansion could broaden a phenotype beyond its intended scope, while
        apparently comprehensive lists could omit codes actually used in the
        source.
  \item \textbf{Sparse data reported without caveat.} Statistics from sparsely
        populated fields can silently describe a subset rather than the cohort,
        and an unrecorded diagnosis could mean unobserved, not absent.
  \item \textbf{Unit, temporal, and endpoint sensitivity.} Mixed units,
        incomplete follow-up, and surrogate endpoints sensitive to windows,
        exclusions, or event ordering could materially change a result.
\end{compactenum}

\noindent Implausible zeros, near-perfect matches, or counts inconsistent with
clinical expectations often indicated the wrong table, vocabulary, unit, or
aggregation level, requiring epidemiological plausibility checks beyond SQL
validation.

\paragraph{Lessons learned.}
First, \emph{feasibility is itself an analytical
outcome}: staged checks of coding, coverage, observability, and follow-up expose
unsupported analyses before substantial effort is invested. Second,
\emph{the errors that matter are silent}: the consequential mistakes we observed
passed execution and returned plausible numbers, so epidemiological plausibility,
not query success, became the effective validation gate. Third, while routine descriptive questions are largely automated (over 95\%
reference-query accuracy~\citep{ziletti-etal-2026-disentangling}), \emph{expert
judgment remains essential for research-level epidemiology}. Practitioners
reported completing within guided sessions tasks that would ordinarily require
days or weeks of SQL development and code-list curation, while retaining
validated code lists, cohorts, and database conventions for reuse.

\section{Conclusion} \label{sec:conclusion}
\textsc{Ascent} exposes governed medical coding, epidemiological analysis, and
cohort generation across OMOP and native schemas through a shared MCP tool
surface, decoupling these capabilities from any particular client, model, or
orchestrator. Capable models in agentic configurations achieve higher accuracy
than the fixed pipeline across three orchestration frameworks but require more
tool calls and time. Real use supports expert-guided analysis that shifts effort
from implementation to feasibility assessment, data-quality auditing, and
diagnostic iteration without replacing scientific judgment.

\section*{Limitations} \label{sec:limitations}
The evaluation uses a single Optum EHR data source and does not cover all
epidemiological tasks. \textsc{EpiTrap} tests one error per item rather than
end-to-end correctness. The fixed pipeline is not compute-matched to the agents,
and grading uses LLMs rather than humans. Usage findings are preliminary and
qualitative. \textsc{Ascent} does not estimate causal effects.

\section*{Acknowledgments}
We thank Amy Mulick, Pareen Vora, Elif Eroglu, and Nasreen Khan for feedback as
system users.

\bibliography{anthology,custom}

\end{document}